\documentclass[10pt,twocolumn,letterpaper]{article}

\usepackage{iccv}              

\newcommand{\FAIL}{{\color{red}\ding{55}}\ }

\newcommand{\boldparagraph}[1]{\vspace{1pt}\noindent{\bf #1} } 

\newcommand{\methodname}{Deblur-SLAM}

\newcommand{\win}[1]{\fs #1}
\newcommand{\cmark}{{\color{PineGreen}\checkmark}}
\newcommand{\xmark}{{\color{red}\ding{55}}\ }

\definecolor{iccvblue}{rgb}{0.21,0.49,0.74}
\usepackage{graphicx} 
\usepackage{amsmath}
\usepackage{amssymb}
\usepackage{bbm}
\usepackage[utf8]{inputenc} 
\usepackage[T1]{fontenc}    
\usepackage{url}            
\usepackage{booktabs}       
\usepackage{amsfonts}       
\usepackage{nicefrac}       
\usepackage{microtype}      
\usepackage{graphicx}
\usepackage{multirow} 
\usepackage{makecell} 
\usepackage{colortbl} 
\usepackage{diagbox} 
\usepackage{siunitx}
\usepackage{enumitem} 
\usepackage{arydshln} 
\usepackage{tabularx}
\usepackage{booktabs} 
\usepackage{xspace} 
\usepackage{pifont} 
\usepackage[dvipsnames]{xcolor} 
\usepackage{dsfont} 
\usepackage{bm}
\usepackage{svg}
\usepackage[pagebackref,breaklinks,colorlinks,allcolors=iccvblue]{hyperref}

\def\ie{\emph{i.e.}\xspace}

\colorlet{colorFst}{Green!25}       
\colorlet{colorSnd}{SpringGreen!45} 
\colorlet{colorTrd}{Yellow!30}      
\colorlet{colorLow}{darkgray!30}    
\newcommand{\fs}{\cellcolor{colorFst}\bf}   
\newcommand{\nd}{\cellcolor{colorSnd}}      
\newcommand{\rd}{\cellcolor{colorTrd}}      

\definecolor{gray}{rgb}{0.65,0.65,0.65}
\definecolor{mycol}{rgb}{0.90,0.95,1.0}

\def\paperID{12513} 
\def\confName{ICCV}
\def\confYear{2025}

\title{Deblur Gaussian Splatting SLAM}

\author{Francesco Girlanda$^{1}$ \qquad  Denys Rozumnyi$^{1}$ \qquad  Marc Pollefeys$^{1,2}$ \qquad  Martin R. Oswald$^{3}$ \\
$^{1}$ETH Z\"urich\qquad
$^{2}$Microsoft\qquad $^{3}$University of Amsterdam \\
}

\begin{document}
\twocolumn[{
\maketitle
\begin{center}
    \vspace{-1.7em}
    \captionsetup{type=figure}
    \includegraphics[width=\textwidth]{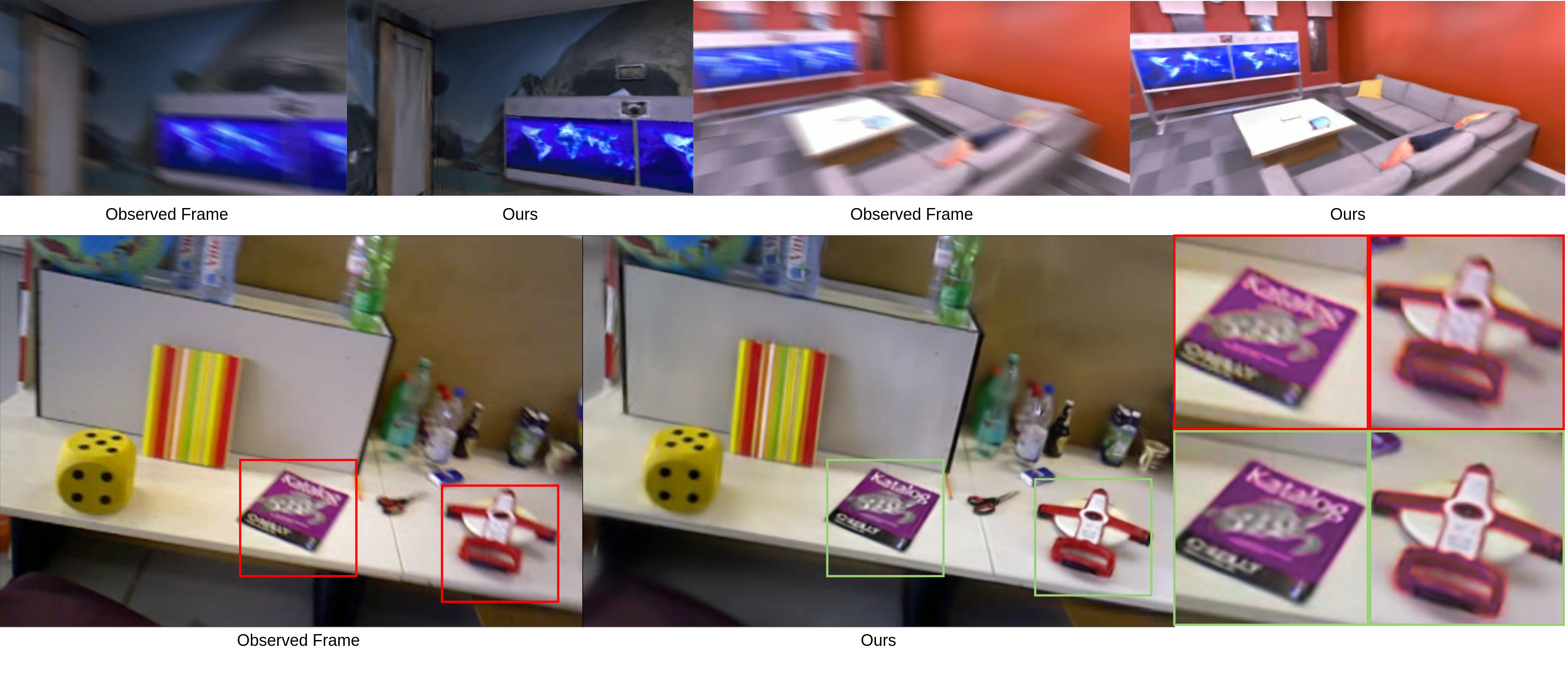}
    \vspace{-2.5em}
    \captionof{figure}{\textbf{\methodname{}} can successfully track the camera and reconstruct sharp maps for highly motion-blurred sequences. We directly model motion blur, which enables us to achieve high-quality reconstructions, both on challenging synthetic (top) and real (bottom) data. \label{fig:preview} }
    \vspace{0.7em}
\end{center}
}]
\begin{abstract}
We present \methodname{}, a robust RGB SLAM pipeline designed to recover sharp reconstructions from motion-blurred inputs. 
The proposed method bridges the strengths of both frame-to-frame and frame-to-model approaches to model sub-frame camera trajectories that lead to high-fidelity reconstructions in motion-blurred settings. 
Moreover, our pipeline incorporates techniques such as online loop closure and global bundle adjustment to achieve a dense and precise global trajectory. 
We model the physical image formation process of motion-blurred images and minimize the error between the observed blurry images and rendered blurry images obtained by averaging sharp virtual sub-frame images. 
Additionally, by utilizing a monocular depth estimator alongside the online deformation of Gaussians, we ensure precise mapping and enhanced image deblurring. 
The proposed SLAM pipeline integrates all these components to improve the results.
We achieve state-of-the-art results for sharp map estimation and sub-frame trajectory recovery both on synthetic and real-world blurry input data. 

\end{abstract}
\section{Introduction}

The challenge of reconstructing the real world with high fidelity has long been a core problem in computer vision, essential for applications such as robotic navigation, VR/AR, architecture, and autonomous vehicles. 
Achieving accurate, detailed representations of physical environments is crucial, and dense visual Simultaneous Localization and Mapping (SLAM) systems address this by estimating a precise reconstruction of an environment while simultaneously localizing the camera frames.

Traditional 3D SLAM approaches typically rely on geometric representations, implemented in various forms, such as weights of an MLP~\cite{azinovic2022neural,Sucar2021IMAP:Real-Time,matsuki2023imode,ortiz2022isdf}, features anchored in dense grids~\cite{zhu2022nice,newcombe2011kinectfusion,Weder2020RoutedFusion,weder2021neuralfusion,sun2021neuralrecon,bovzivc2021transformerfusion,li2022bnv,zou2022mononeuralfusion,uncleslam2023}, hierarchical octrees~\cite{yang2022vox} and points/surfels~\cite{keller2013real,schops2019bad,teed2021droid}, mesh~\cite{bloesch2019learning}, and voxel hashing~\cite{zhang2023go,zhang2023hislam,chung2022orbeez,Rosinol2022NeRF-SLAM:Fields,matsuki2023newton}. 
Recent visual SLAM approaches further capture visual appearance, drawing on advances in Neural Radiance Fields (NeRF)~\cite{Mildenhall2020NeRF:Synthesis} and its variants~\cite{kerbl3Dgaussians,muller2022instant}, allowing for photorealistic image synthesis of environments.
These advances enable new possibilities in complex downstream tasks, including detailed semantic scene understanding~\cite{jatavallabhula2023concept}, language-guided manipulation~\cite{shen2023distilled}, and visual navigation~\cite{shafiullah2023clipfields}. Additionally, neural representations have the advantage of filling unseen regions with smooth geometric estimation and offering a low-memory footprint~\cite{ortiz2022isdf,Sucar2021IMAP:Real-Time,wang2023coslamjointcoordinatesparse}.

Methods based on 3D Gaussian Splatting (3DGS) ~\cite{kerbl3Dgaussians} yield high-fidelity renderings~\cite{matsuki2024gaussian, yugay2023gaussianslam, keetha2024splatam, huang2023photo, yan2023gs,sandstrom2024splat} and show promising results in the RGB-only setting, given their flexibility in optimizing surface locations~\cite{matsuki2024gaussian,huang2023photo,sandstrom2024splat}. 
However, these methods often lack multi-view depth and geometric priors, leading to suboptimal geometry in RGB-only applications. 
Most approaches also only implement frame-to-model tracking without global trajectory and map optimization, which can result in excessive drift, especially under real-world conditions. 
Frame-to-frame tracking methods, combined with loop closure and global bundle adjustment (BA), still achieve state-of-the-art tracking accuracy~\cite{zhang2023hi, zhang2023go,zhang2024glorie}. 
The current state-of-the-art in SLAM is SplatSLAM~\cite{sandstrom2024splat}, which seeks to incorporate the strengths of frame-to-frame methods by integrating loop closure, global bundle adjustment, and deformable 3D Gaussian maps. 

At the same time, motion blur is a common phenomenon that degrades image quality due to fast camera motion -- a common failure of most state-of-the-art SLAM and 3D reconstruction methods.
These methods do not employ any strategy to handle severe motion-blurred images that usually lead to worse reconstruction quality.

The proposed \methodname{} method aims to solve the under-constrained motion-blurred setting.
It combines both frame-to-frame and frame-to-model approaches and estimates sub-frame trajectories that lead to sharper reconstructions (see Fig.~\ref{fig:preview}). 
We make the following \textbf{contributions}:
\begin{itemize}
\item We propose an RGB SLAM method that explicitly addresses and models camera motion blur. This leads to sharp maps even when input frames are highly blurred by fast camera motion.
\item For increased robustness, we combine frame-to-frame and frame-to-model trackers and, for the first time, incorporate loop closure, bundle adjustment, and online refinement into a deblurring SLAM framework.
\item In contrast to most SLAM methods, we estimate trajectories with sub-frame precision.
\end{itemize}

\section{Related work}
\boldparagraph{Dense visual SLAM.}
Curless and Levoy~\cite{curless1996volumetric} pioneered dense online 3D mapping using truncated signed distance functions (TSDF). KinectFusion~\cite{newcombe2011kinectfusion} demonstrated real-time SLAM through depth maps. 
To address pose drift, globally consistent pose estimation and dense mapping techniques have been developed, often by dividing the global map into submaps~\cite{cao2018real,dai2017bundlefusion,fioraio2015large,tang2023mips,matsuki2023newton,maier2017efficient,kahler2016real,stuckler2014multi,choi2015robust,Kahler2015infiniTAM,reijgwart2019voxgraph,henry2013patch,bosse2003atlas,maier2014submap,tang2023mips,mao2023ngel,liso2024loopyslam}. 
Loop detection triggers submap deformation via pose graph optimization~\cite{cao2018real,maier2017efficient,tang2023mips,matsuki2023newton,kahler2016real,endres2012evaluation,engel2014lsd,kerl2013dense,choi2015robust,henry2012rgb,yan2017dense,schops2019bad,reijgwart2019voxgraph,henry2013patch,stuckler2014multi,wang2016online,hu2023cp,mao2023ngel,liso2024loopyslam}, and global bundle adjustment (BA) is employed for refinement~\cite{dai2017bundlefusion,schops2019bad,cao2018real,teed2021droid,yan2017dense,yang2022fd,matsuki2023newton,chung2022orbeez,tang2023mips,hu2023cp}. 
While 3D Gaussian SLAM with RGB-D input has demonstrated promising results, these methods do not consider global consistency through mechanisms like loop closure~\cite{yugay2023gaussianslam,keetha2024splatam,zhang2023go}. 
Point-SLAM~\cite{sandstrom2023point} creates photorealistic 3D maps using a dynamic neural point cloud.
DROID-SLAM~\cite{teed2021droid} achieves global consistency directly minimizing reprojection errors, which iteratively refines dense optical flow and camera poses. 
Recent advancements like GO-SLAM~\cite{zhang2023go}, HI-SLAM~\cite{zhang2023hislam}, and GlORIE-SLAM~\cite{zhang2024glorie} optimize factor graphs for accurate tracking. 
For a recent in-depth survey on NeRF-inspired dense SLAM, we refer to~\cite{tosi2024nerfs}.

\boldparagraph{RGB SLAM.}
Most research concentrates on combining RGB and depth cameras. 
RGB-only methods are less explored due to the lack of geometric data, which introduces scale, depth ambiguities, motion blur and motion parallax effects. Nonetheless, dense SLAM using only RGB cameras is attractive for its cost-effectiveness and versatility across different environments.
MonoGS~\cite{matsuki2024gaussian} and Photo-SLAM~\cite{huang2023photo} were pioneers in implementing RGB-only SLAM using 3D Gaussians. However, their lack of proxy depth information limits their ability to achieve high-accuracy mapping. 
MonoGS~\cite{matsuki2024gaussian} does not ensure global consistency either. 
SplatSLAM~\cite{sandstrom2024splat} incorporates global consistency but lacks strategies to handle motion blur as well as modeling dense sub-frame trajectories.

\boldparagraph{Image deblurring.}
Motion blur significantly impacts the output quality of many computer vision methods and has been a longstanding subject of active research. 
Various studies have addressed the effects of motion blur, enabling the reconstruction of sharp 3D maps when the input images are blurred~\cite{dai2023hybrid, ma2022deblur,lee2024deblurring,darmon2024robust,guo2021deblurslam,seiskari2024gaussiansplattingmoveblur}. 
In visual SLAM systems, input measurements are particularly susceptible to motion blur because a moving camera captures images over the exposure period. 
To mitigate this, we parametrize the camera trajectory during the exposure time and optimize sub-frame camera poses to match the blurred inputs, similar to the approaches in~\cite{sfb,mfb,hfb,wang2023bad,zhao2025bad,bae20242}. 
Similar to our setting, $I^2$-SLAM~\cite{bae20242} aims to reconstruct photorealistic and sharp maps from casually captured videos that contain severe motion blur and varying appearances. 
In contrast to our work, it lacks global consistency through loop closure, bundle adjustment, an online refinement of the deformable 3D Gaussian map, and, specifically for the RGB setting, a monocular depth estimator to improve Gaussian initialization.

\begin{figure*}
\centering
\includegraphics[width=\linewidth]{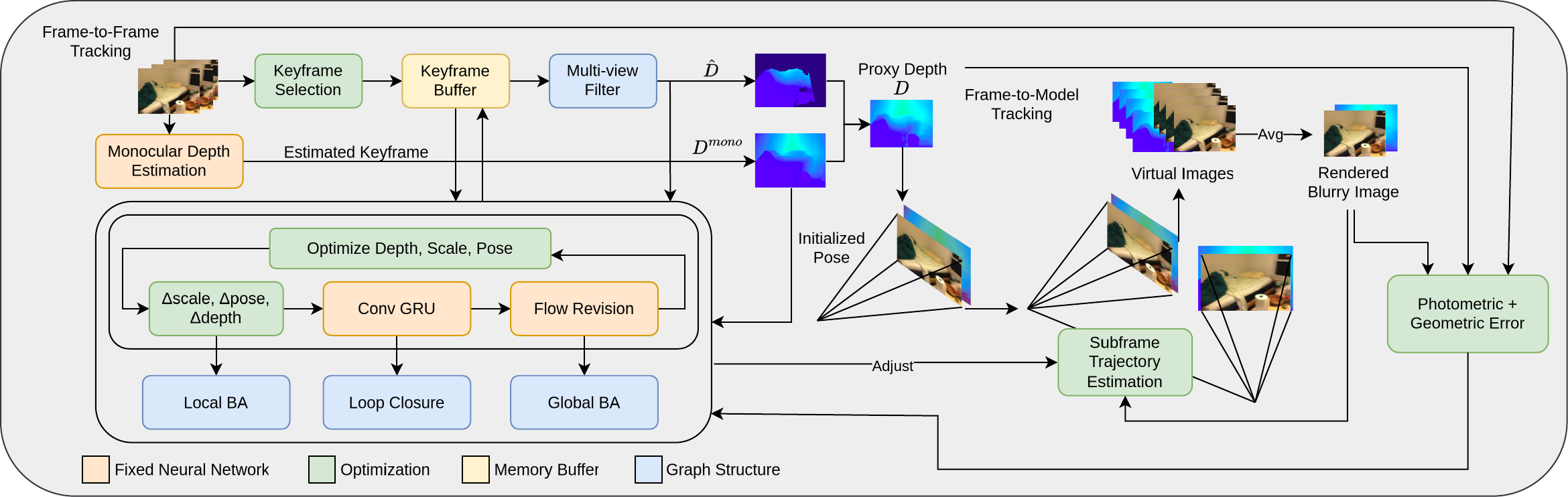}
\caption{\textbf{Architecture of \methodname{}.} Given an RGB input stream, we estimate an initial pose through local bundle adjustment (BA) using joint Disparity, Scale and Pose Optimization (DSPO). 
This pose is later refined through frame-to-model tracking that learns a sub-frame trajectory. Each keyframe is then mapped, taking advantage of the estimated monocular depth. 
The sub-frame trajectory is applied to render virtual sharp images, which model the physical image formation of blurry images. 
We optimize the photometric and geometric error between the observed blurry image and the average of our sharp images.
We further refine poses globally via online loop closure, global BA, and a deformable 3D Gaussian map that adjusts for global pose and depth updates before each mapping phase.
}
\label{fig:architecture}
\end{figure*}

The proposed \methodname{} overcomes these limitations by leveraging a globally consistent tracker, a monocular depth estimator, and a deformable 3D Gaussian map that adapts online to loop closure and global bundle adjustment, as in~\cite{sandstrom2024splat}. 
By directly modeling the camera motion blur, we present a complete pipeline that addresses blur, which is the most common issue of state-of-the-art SLAM methods. 

%

%
\section{Method}
\label{sec:method}
\methodname{} aims to bridge the strengths of both frame-to-frame and frame-to-model approaches in order to get sub-frames trajectories that lead to sharp reconstructions in a motion-blurred setting. 
We leverage SplatSLAM~\cite{sandstrom2024splat}’s frame-to-frame features to provide precise initialization of both camera poses and Gaussians.
Then, we stack a frame-to-model tracker on top to learn the sub-frame trajectory, allowing sharp images to be rendered at sub-frame precision. 
We refer to these sharp images as ‚virtual images‘, which can be used to generate a video with a higher frame rate.  
We optimize the photometric and geometric error between the observed blurry image and the synthesized blurred image, obtained by averaging the sharp virtual sub-frames (Fig.~\ref{fig:architecture}). 
During optimization, we use an adjustment process to regularize the sub-frame trajectory, aligning it with the global trajectory to achieve a highly accurate and detailed reconstruction. 
These three components -- frame-to-frame tracker, frame-to-model tracker, and blur decomposition -- are the main components of the proposed method, which are detailed in the following subsections.

%
\begin{figure*}
\centering
  \includegraphics[width=\linewidth]{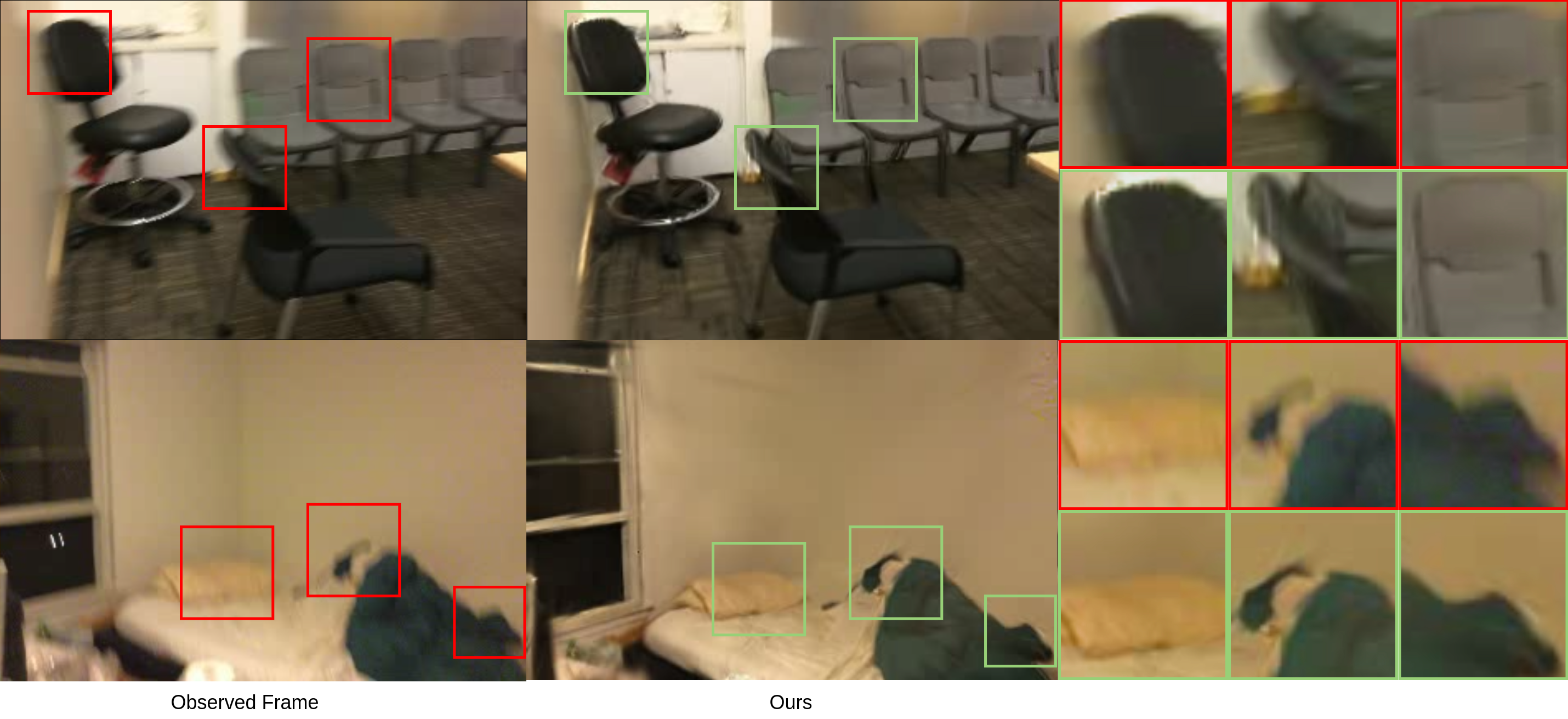}\\
  \vspace{-0.3em}
\caption{\textbf{Qualitative results on real-world ScanNet~\cite{Dai2017ScanNet} data.} Given the input blurry frames from scenes 0169 and 0207, we manage to track the trajectory with sub-frame precision and estimate sharp maps by directly modeling the camera motion blur.} 
\label{fig:scannet_qual_169_207}
\end{figure*}

\subsection{Motion blur image formation model}

The physical formation process of an image in a digital camera consists of gathering photons during the exposure period that are later converted to a digital image. 
This process can be formalized by integrating a sequence of sharp images:
\begin{equation}
    B(u) = \phi \int_0^{\tau} C_t(u) dt \enspace ,
\end{equation}
where $B(u) \in \mathbb{R}^{3}$ represents the motion-blurred image, $u \in \mathbb{R}^2$ is a pixel in the image, $\phi$ is a normalization factor, $\tau$ is the exposure time, $C_t(u) \in \mathbb{R}^{3}$ is the virtual sharp image captured at sub-frame time $t \in \left[ 0, \tau \right]$.
This camera model approximates motion blur by discretizing into $M$ timestamps (or virtual views):
\begin{equation}
    B(u) \approx \frac{1}{M} \sum_{i=0}^{M-1} C_i(u) \enspace .
\end{equation}
The camera trajectory representation is parametric and has a fixed number of control points (by default, we use 2). 
We use $M$ virtual cameras/views that are obtained through a linear interpolation between the control points using LERP (Linear intERPolation) for the translation vector and SLERP (Spherical Linear intERPolation) for the rotation matrix. 
By rendering views from the virtual cameras, we obtain so-called virtual sharp images.
The pipeline can be easily extended to more complex interpolations, such as the cubic B-spline interpolation with four control points in the \textbf{SE(3)} space. 
We refer to prior work \cite{liu2021mba, wang2023bad, zhao2025bad} for details about the interpolation and derivations of the related Jacobian.

\subsection{Deformable 3D Gaussian Splatting}
We adopt a deformable 3D Gaussian Splatting representation that allows us to achieve global consistency.
The scene is represented by a set $\mathcal{G} = \{g_i\}_{i=1}^{N}$ of $N$ 3D Gaussians. Each Gaussian $g_i$ is parameterized by its centroid $\mu_i \in \mathbb{R}^3$, 3D covariance $\Sigma \in \mathbb{R}^{3 \times 3}$, opacity $\sigma \in \left[0, 1\right]$ and color $c \in \mathbb{R}^3$. The distribution of each scaled Gaussian is defined as:
\begin{equation}
g_i(\mathbf{x})=\exp \left(-\frac{1}{2}\left(\mathbf{x}-\boldsymbol{\mu}_i\right)^{\top} \Sigma_i^{-1}\left(\mathbf{x}-\boldsymbol{\mu}_i\right)\right) \, .
\end{equation}
As in previous methods, we ensure that the 3D covariance $\Sigma$ remains positive semi-definite. 
To reduce the optimization complexity, 3DGS represents $\Sigma_i$ using a scale $S_i \in \mathbb{R}^3$ and rotation matrix $R_i \in \mathbb{R}^{3 \times 3}$ stored by a quaternion $q_i \in \mathbb{R}^4$ leading to the decomposition $\Sigma_i = R_i S_i S_i^T R_i^T$.

\boldparagraph{Rendering.}
To render 3D Gaussians to a 2D plane, we project the covariance matrix $\Sigma$ as $\Sigma' = J R \Sigma R^T J^T$ and project $\mu$ as $\mu' = K \omega^{-1} \mu$, where $R$ is the rotation component of world-to-camera extrinsic $\omega^{-1}$, and $J$ is the Jacobian of the affine approximation of the projective transformation~\cite{zwicker2001surface}. 
The final pixel color $C$ and depth $D^r$ at pixel $x'$ is rendered by rasterizing the obtained Gaussians that overlap with the given pixel, sorted by their depth as
\begin{equation}
C=\!\sum_{i \in \mathcal{N}} \mathbf{c}_i \alpha_i \prod_{j=1}^{i-1}\!\left(1\!-\!\alpha_j\right) \, , \;\;\; D^r=\!\sum_{i \in \mathcal{N}} \hat{d}_i \alpha_i \prod_{j=1}^{i-1}\!\left(1\!-\!\alpha_j\right)
\end{equation}
where $\hat{d}_i$ is the z-axis depth of the center of the $i$-th Gaussian. 
We obtain the 2D opacity $\alpha_i$ by multiplying the 3D opacity $o_i$ and the 2D Gaussian density as
\begin{equation}
\alpha_i=o_i \exp \left(-\frac{1}{2}\left(\mathbf{x}^{\prime}-\boldsymbol{\mu}_i^{\prime}\right)^{\top} \Sigma_i^{\prime-1}\left(\mathbf{x}^{\prime}-\boldsymbol{\mu}_i^{\prime}\right)\right) \, .
\end{equation}

\boldparagraph{Map initialization.}
We assume the camera remains static for at least one initial frame at the start of the sequence. 
This approach ensures a sharp frame at the beginning of our pipeline, providing a reliable basis for effectively initializing the Gaussians.
This assumption is common in real-life use cases and hold true for all tested real-world datasets.
To improve the map initialization in the RGB setting, we make use of a proxy depth map $D$ that combines the inlier multi-view depth $\tilde{D}$ and the monocular depth $D^{\text{mono}}$:
\begin{equation}
D(u, v)= \begin{cases}\tilde{D}(u, v) & \text { if } \tilde{D}(u, v) \text { is valid } \\ \theta D^{\text {mono }}(u, v)+\gamma & \text { otherwise }\end{cases} ,
\end{equation}
where $\theta$ and $\gamma$ are computed as in the initial least square fitting (see supplementary material).

\begin{figure*}
\centering
 \includegraphics[width=\linewidth]{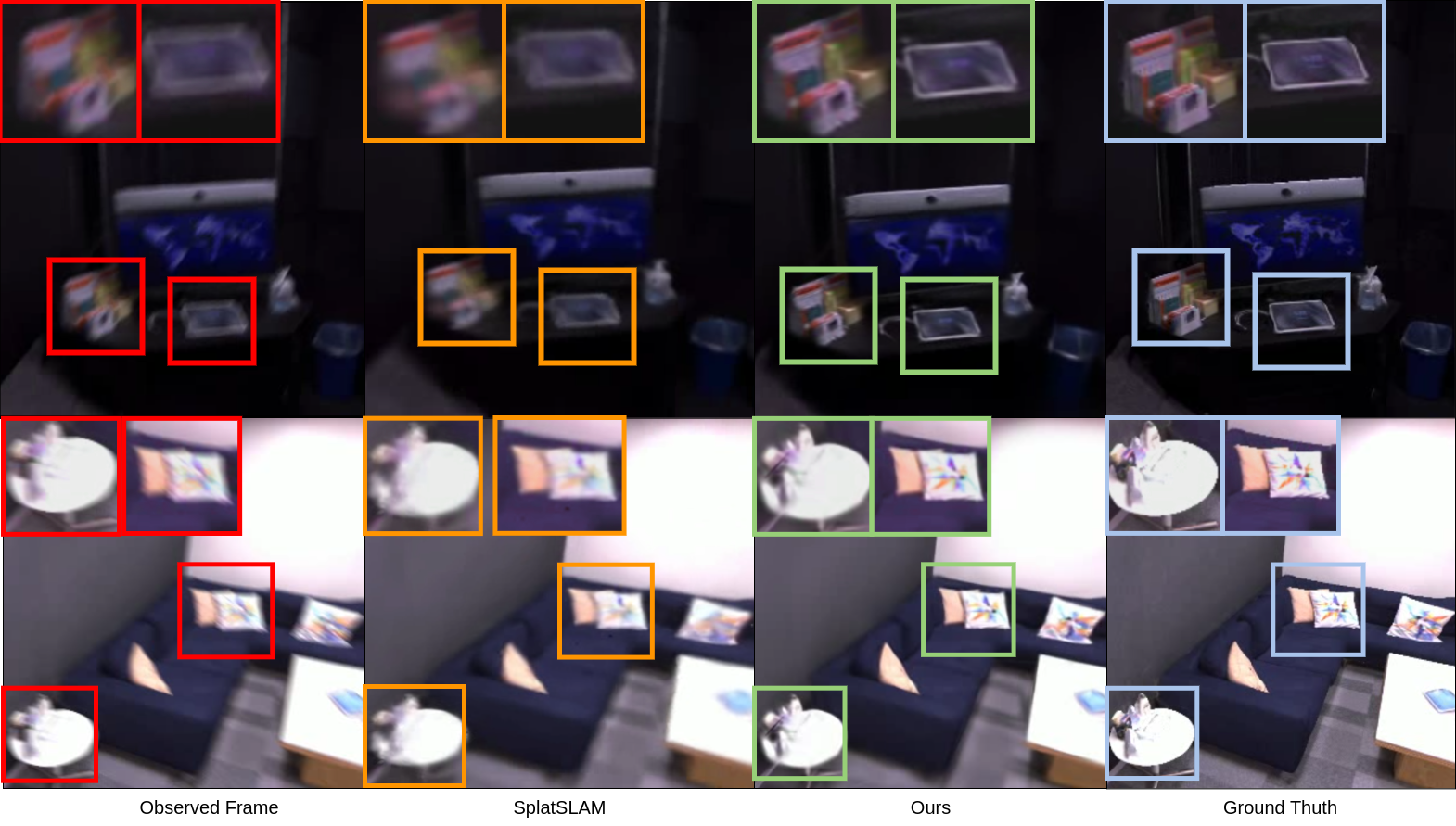}\\ 
 \vspace{-0.4em}
\caption{\textbf{Qualitative comparison to Splat-SLAM.} We compare in both a darker scene (office 1 -- top) and a very bright scene (office 2 -- bottom) on the newly proposed synthetically blurred Replica dataset~\cite{straub2019replica}, showing the versatility of the proposed model. 
From left to right: the observed frame with visible camera motion blur, the SplatSLAM~\cite{sandstrom2024splat} rendering, one of our sharp virtual images, and the ground truth. 
The proposed \methodname{} method can recover more details such as the boxes (top) and the pillow (bottom).} 
\label{fig:office2_qual}
\end{figure*}

\boldparagraph{Keyframe selection.} 
We employ the strategies from \cite{matsuki2024gaussian, sandstrom2024splat}, \ie we use both an optical flow threshold $\tau_f$ and a keyframe selection strategy to avoid mapping redundant frames. 
Since our viewpoints now consist of interpolating two control points, we calculate the intersection over union (IoU) on the viewpoint using the midpoint between them.

\boldparagraph{Mapping loss.} 
The rendered blurry image $B_k$ is defined as:
\begin{equation}
\label{eq:blurry_img_formation}
    B_k = a_k \left( \frac{1}{M} \sum_{i=0}^{M-1} C_i \right) + b_k \enspace,
\end{equation}
where $a_k$ and $b_k$ are optimizable parameters to account for varying exposure and lightning changes, and $M$ is the number of virtual images.

We optimize the difference between $B_k$ and the observed input image, applying a photometric and geometric loss to the image and a scale regularizer to avoid artifacts from elongated Gaussians. 
Our photometric loss is defined as:
\begin{align}
    \mathcal{L}_\text{ph} 
    &= \lambda_\text{SSIM} \mathcal{L}_\text{SSIM} + (1 - \lambda_\text{SSIM} ) \left|B_k-B_k^{g t}\right|_1 \, ,
\end{align}
where $B_k$ is the image after the exposure correction of equation \eqref{eq:blurry_img_formation}. 
This term, together with the geometric loss and the scale regularizer, leads to the final optimization objective:
\begin{align}
 \min _{\mathcal{G}, \mathbf{a}, \mathbf{b}} \sum_{k\!\in\!\text{KFs}} \!\frac{\lambda}{N_k} \mathcal{L}_\text{ph} \!+\! \frac{1\!-\!\lambda}{N_k}\left|\bar{D}_k^r-D_k\right|_1 
 \!+\! \frac{\lambda_\text{reg}}{|\mathcal{G}|} \!\sum_i^{|\mathcal{G}|}\left|s_i-\tilde{s}_i\right|_1 \, ,
\end{align}
where KFs contains the set of keyframes in the local window, $N_k$ is the number of pixels per keyframe, $\bar{D}_k^r$ is the average depth of the virtual views at frame $k$.
We denote hyperparameters as $\lambda$, $\lambda_\text{SSIM}$ and $\lambda_\text{reg}$.
The mean scaling is represented by $\tilde{s}$. 
After the end of the online process, we perform a set of final refinements as in \cite{matsuki2024gaussian,zhang2024glorie,sandstrom2024splat}.

%
\boldparagraph{Map deformation.}
Since we update keyframe poses and proxy depth maps online, the 3D Gaussian map is adjusted by a non-rigid deformation as in \cite{sandstrom2024splat}.  
We update the mean, scale, and rotation of all Gaussians $g_i$ that are associated with the given keyframe. 
The mean $\mu_i$ is projected into $\omega$ to find pixel correspondence $(u, v)$. 
Since the Gaussians are not necessarily anchored on the surface, instead of re-anchoring the mean at multi-view depth $D'$, we shift the mean by $D'(u, v) - D(u, v)$ along the optical axis and update $R_i$ and $s_i$ accordingly as
\begin{align}
  \boldsymbol{\mu}_i^{\prime} &=\left(1+\frac{D^{\prime}(u, v)-D(u, v)}{\left(\omega^{-1} \boldsymbol{\mu}_i\right)_z}\right) \omega^{\prime} \omega^{-1} \boldsymbol{\mu}_i \, , \\
  R_i^{\prime}&=R^{\prime} R^{-1} R_i, s_i^{\prime}=\!\left(\!1+\frac{D^{\prime}(u, v)-D(u, v)}{\left(\omega^{-1} \boldsymbol{\mu}_i\right)_z}\!\right)\! s_i \, .
\end{align}

\subsection{Frame-to-frame tracking}
In the tracking phase, we initialize the pose using a pre-trained recurrent optical flow model coupled with a Disparity, Scale and Pose Optimization (DSPO) objective from~\cite{teed2021droid,sandstrom2024splat}, to optimize camera poses jointly with per-pixel disparities.

The optimization process uses the Gauss-Newton algorithm on a factor graph $G(V,E)$, where the nodes $V$ represent the keyframe poses and disparities, and edges $E$ model the optical flow between these keyframes. 
Odometry keyframe edges are incorporated into $G$ by calculating the optical flow relative to the previously added keyframe. 
If the mean flow exceeds a threshold $\tau_f \in \mathbb{R}$, a new keyframe is introduced into $G$. The DSPO objective contains two distinct optimization objectives that are addressed in an alternating fashion. The primary objective, known as Dense Bundle Adjustment (DBA)~\cite{Tang2018BANetDB,teed2021droid}, jointly optimizes the pose and disparity of the keyframes, as shown in equation \eqref{DBA}. 
This objective operates on a sliding temporal window centered around the current frame, \ie
\begin{equation}
  \label{DBA}
  \underset{\omega, d}{\arg \min }\!\!\! \sum_{(i, j) \in E} \! \left\|\tilde{p}_{i j}-K \omega_j^{-1}\!\left(\omega_i\left(1 / d_i\right) K^{-1}\!\left[p_i, 1\right]^T\right)\!\right\|_{\Sigma_{i j}}^2 \, ,
\end{equation}
where $\tilde{p}_{ij} \in \mathbb{R}^{(W \times H \times 2) \times 1}$ represents the predicted pixel coordinates after warping $p_i$ into keyframe $j$ via optical flow. 
Matrix $K$ denotes the camera intrinsics, while $\omega_j$ and $\omega_i$ denote camera-to-world extrinsics. The Mahalanobis distance $\|\cdot\|_{\Sigma_{ij}}$ incorporates confidence-based weights.

The second objective, called Disparity, Scale and Pose Optimization (DSPO), refines disparity using monocular depth $D^\text{mono}$ from a pretrained DPT model~\cite{Ranftl2020,Ranftl2021}, ensuring correct scale and shift estimation:
\begin{align}
  \underset{d^h, \theta, \gamma}{\arg \min }& \sum_{(i, j) \in E}\left\|\tilde{p}_{i j}-K \omega_j^{-1}\left(\omega_i\left(1 / d_i^h\right) K^{-1}\left[p_i, 1\right]^T\right)\right\|_{\Sigma_{i j}}^2 \nonumber \\
  + & \alpha_1 \sum_{i \in V}\left\|d_i^h-\left(\theta_i\left(1 / D_i^{\text{mono }}\right)+\gamma_i\right)\right\|^2 \\
  + & \alpha_2 \sum_{i \in V}\left\|d_i^l-\left(\theta_i\left(1 / D_i^{\text{mono }}\right)+\gamma_i\right)\right\|^2. \nonumber
\end{align}
Here, the optimizable parameters are scales $\theta \in \mathbb{R}$, shifts $\gamma \in \mathbb{R}$ and a subset of disparities $d^{h}$ classified as being high error (see supplementary material). 
This approach is used as the monocular depth is only valuable in regions where the multi-view disparity $d_i$ optimization lacks accuracy. 
Additionally, setting $\alpha_1 < \alpha_2$ ensures that scales $\theta$ and shifts $\gamma$ are optimized with the preserved low-error disparities $d^{l}$. 
The scale $\theta_i$ and shift $\gamma_i$ are initially estimated using least squares fitting (see supplementary material). 
By optimizing DBA and DSPO alternatingly, we avoid the scale ambiguity encountered if $d$, $\theta$, $\gamma$, and $\omega$ are optimized jointly.

\begin{table*}
\centering
\scriptsize
\setlength{\tabcolsep}{6.5pt}  
\begin{tabularx}{\linewidth}{llrrrrrrrrrr}
\toprule
Method & Metric & \texttt{office0} & \texttt{office1} & \texttt{office2} & \texttt{office3} & \texttt{office4} & \texttt{room0} & \texttt{room1} & \texttt{room2} & Avg. & Avg. w.o. \FAIL \\
\midrule

\multirow{4}{*}{\makecell[l]{MonoGS~\cite{matsuki2024gaussian}}} 
& sPSNR$\uparrow$ &   \rd 25.30 & \rd  27.82 &  \rd 21.81 & \rd  23.91 &  \rd 23.61 & \rd  22.21 &  \nd 20.95 &  \nd 24.95 & \nd  23.82 & \rd 24.11 \\
& sSSIM$\uparrow$ & \rd  0.82 & \win{0.88} & \nd  0.81 & \nd  0.83 & \nd  0.85 & \rd  0.72 &  \nd 0.74 &  \nd 0.82 & \nd  0.81 & \nd 0.82 \\
& sLPIPS$\downarrow$ &  \rd 0.43 & \nd  0.30 & \rd  0.40 & \rd  0.32 & \rd  0.40 & \rd  0.47 &  \nd 0.48 & \nd  0.37 &  \nd 0.40 & \rd 0.39 \\
& sATE[cm]$\downarrow$ &  \rd  30.84 & \rd  13.97 & \rd  50.16 & \rd  3.69 & \rd  108.62 & \rd  44.05 & \nd  51.80 & \nd  11.51 & \nd  39.33 & \rd 41.89\\
[0.8pt] \hdashline \noalign{\vskip 1pt}

\multirow{4}{*}{\makecell[l]{Splat-\\SLAM~\cite{sandstrom2024splat}}} 
& sPSNR$\uparrow$ & \nd  28.52 &  \nd 28.88 & \nd  22.53 &  \nd 25.01 &  \nd 25.25 &  \nd 23.71 & \rd \FAIL & \rd \FAIL & \rd \FAIL & \nd 25.65 \\
& sSSIM$\uparrow$ & \nd  0.84 & \nd  0.87 & \rd 0.79 & \rd  0.81 & \rd  0.84 & \nd  0.74 & \rd \FAIL & \rd \FAIL & \rd \FAIL & \nd 0.82 \\
& sLPIPS$\downarrow$ & \nd  0.36 & \nd  0.30 &  \nd 0.35 & \nd  0.27 &  \nd 0.34 & \nd  0.39 & \rd \FAIL & \rd \FAIL & \rd \FAIL & \nd 0.34 \\
& sATE[cm]$\downarrow$ & \win{2.88} & \win{2.12} & \nd  2.34 & \nd  2.05 &  \nd 4.53 & \nd  1.90 & \rd \FAIL & \rd \FAIL & \rd \FAIL & \nd  2.64\\
[0.8pt] \hdashline \noalign{\vskip 1pt}

\multirow{4}{*}{\makecell[l]{\textbf{\methodname{}}\\ \textbf{(Ours)}}} 
& sPSNR$\uparrow$ & \win{30.40} & \win{30.29} & \win{25.20} & \win{26.52} & \win{27.06} & \win{25.75} & \win{25.25} & \win{27.74} & \win{27.28} & \win{27.54} \\
& sSSIM$\uparrow$ & \win{0.87} & \win{0.88} & \win{0.84} & \win{0.87} & \win{0.86} & \win{0.80} & \win{0.76} & \win{0.85} & \win{0.84} & \win{0.85} \\
& sLPIPS$\downarrow$ & \win{0.24} & \win{0.21} & \win{0.27} & \win{0.16} & \win{0.25} & \win{0.23} & \win{0.34} & \win{0.22} & \win{0.24} & \win{0.23} \\
& sATE[cm]$\downarrow$ & \nd 3.04 & \nd  2.94 & \win{1.52} & \win{1.00} & \win{3.75} & \win{1.08} & \win{5.87} & \win{3.78} & \win{2.87} & \win{2.22} \\
\bottomrule
\end{tabularx}
\caption{\textbf{Sub-frame evaluation on Replica~\cite{straub2019replica}.} Since this is a synthetic dataset, ground truth sub-frame trajectories and renderings are available. 
Thus, we evaluate the sub-frame version of PSNR, SSIM, LPIPS, and ATE (denoted by "s" in front). Our method achieves new state-of-the-art performance in estimating those sub-frame trajectories and renderings. A complete failure of the method is denoted by \FAIL. The best-performing method is denoted by bold font and dark green color, second best by light green, and worst-performing by yellow. }
\label{tab:render_replica}
\end{table*}

\boldparagraph{Loop closure.}
To reduce scale and pose drift, we incorporate loop closure and online global bundle adjustment (BA) alongside local window frame tracking. As in \cite{sandstrom2024splat}, loop detection is performed by calculating the mean optical flow magnitude between the currently active keyframes within the local window and all previous keyframes. 
For each keyframe pair, we assess that the optical flow must be below a specified threshold $\tau_{loop}$, and that the time interval between frames must exceed a set threshold $\tau_t$ ensuring co-visibility between the views. When both criteria are satisfied, a unidirectional edge is introduced into the graph. 

\begin{figure}
\centering
\includegraphics[width=\linewidth]{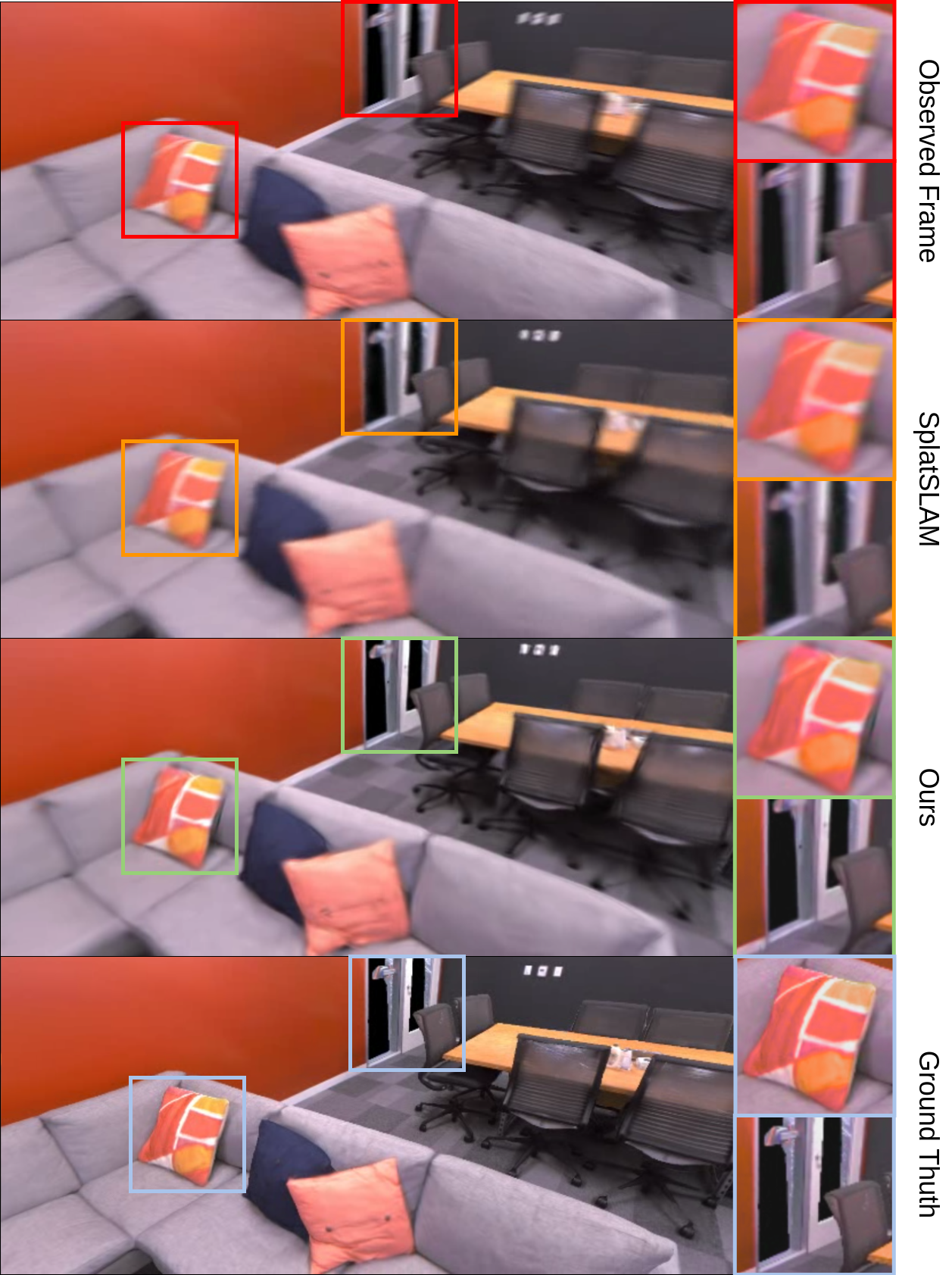}
\caption{\textbf{Qualitative results on the Replica dataset (office 3 scene)~\cite{straub2019replica}.} We outperform Splat-SLAM~\cite{sandstrom2024splat} on blurry data.} 
\label{fig:office3_qual}
\end{figure}

\boldparagraph{Global BA.}
For the online global bundle adjustment (BA), a separate graph is constructed to include all keyframes up to the current point. Edges in this graph are added on the basis of the temporal and spatial relationships between keyframes. 
To ensure numerical stability, the scales of the disparities and poses are normalized before each global BA optimization. 

\subsection{Frame-to-model tracking}
In contrast to Splat-SLAM~\cite{sandstrom2024splat}, we further refine the previous frame-to-frame tracking by frame-to-model tracking using our deblur image formation model.
Using the frame-to-frame pose as the initial pose (treated as a pose in the middle of the sub-frame trajectory), we run iterative optimization on the pose and affine brightness parameters for varying exposure.
This initialization helps us to align with the global trajectory, which is estimated by the DSPO objective. 
This step also allows us to minimize the difference between the average of generated sharp virtual subframes and the observed blurry image.
The optimization objective is as follows:
\begin{equation}
\mathcal{L}_1 = \frac{1}{N} \sum_{i=1}^{N} \Big( \overline{\alpha}(i) \cdot \big| B(i) \cdot W(i) - B_{\text{gt}}(i) \cdot W(i) \big| \Big) \enspace ,
\end{equation}
where $\overline{\alpha}$ represents the mean of opacities that allows us to mask the yet unmapped part of the input image.
Then, $W$ is an edge mask used to calculate the gradient only on the edge pixel, $B(i)$ is a pixel of the image after the exposure correction as defined in \eqref{eq:blurry_img_formation}, and $B_{gt}$ is the blurry observed image.
$N$ is the number of pixels.
Frame index $k$ is omitted for simplicity.

While tracking further frames, we take advantage of the frame-to-frame online optimization and adjust the sub-frame trajectories using linear interpolation to follow the learned global trajectory.

\boldparagraph{Exposure gap estimation.}
In the above optimization, we further introduce two parameters to model the exposure gap between input frames. This helps improving the rendering quality on real-world datasets. We learn these two parameters by rendering each keyframe’s previous and next frames following the same procedure used for the frame-to-model step. 
In addition, we include a regularization term to ensure that the learned subframe trajectory remains smooth. Concretely, we minimize the distance between the end of the previous frame and the beginning of the current keyframe, taking the gap parameter into account. The same approach is applied between the current keyframe and the next frame.

To achieve this, we compute two interpolation points, one between the previous and the current frame, as well as one between the current and the next frame:
\begin{equation}
t_{\mathrm{prev}} = 0.5 + g_\mathrm{prev}
\; ,\quad
t_{\mathrm{next}} = 0.5 + g_\mathrm{next},
\end{equation}
which indicates how far along the motion is between the midpoint extrinsics of the respective frames. $g_\mathrm{prev}, g_\mathrm{next}$ represent the exposure gaps. We use linear interpolation for the translation components and SLERP for the quaternions to obtain the expected start and end pose. We then compare this interpolated pose with the actual viewpoint pose by measuring the squared norm of their translational difference and the squared quaternion angle difference of their rotations:
\begin{align}
    \mathcal{L}_{\text{trans}} 
    &= \bigl\|\mathbf{t}_{\text{viewpoint}} \;-\; \mathbf{t}_{\text{interp}}\bigr\|^2,
    \label{eq:trans_loss} \\
    \mathcal{L}_{\text{rot}} 
    &= \bigl\|2 \,\arccos\Bigl(\bigl|\mathbf{q}_1^{-1} \cdot \mathbf{q}_2\bigr|\Bigr)\bigr\|^2.
    \label{eq:rot_loss}
\end{align}
where $\mathcal{L}_{\text{trans}}$ is the squared norm of the difference between the translation vectors, and $\mathcal{L}_{\text{rot}}$ is the squared norm of the angular difference between the two quaternions corresponding to the rotation matrices.

\section{Experiments}

\begin{table*}
\centering
\scriptsize
\setlength{\tabcolsep}{15.7pt}
\renewcommand{\arraystretch}{1.1}
\begin{tabularx}{\linewidth}{llrrrrrrr}
\toprule
Method & Metric & \texttt{0000} & \texttt{0059} & \texttt{0106} & \texttt{0169} & \texttt{0181} & \texttt{0207} & \bf Avg.\\
\midrule

\multirow{1}{*}{\makecell[l]{MonoGS~\cite{matsuki2024gaussian}}} 
& ATE [cm]$\downarrow$ &  \rd 149.2 &  \rd 96.8 & \rd  155.5 & \rd  140.3 &  \rd 92.6 & \rd  101.9 & \rd  122.7\\
[0.8pt] 

\multirow{1}{*}{\makecell[l]{Splat-SLAM~\cite{sandstrom2024splat}}} 
& ATE [cm]$\downarrow$ & \win{5.5} & \nd 9.1 & \win{7.0} &  \nd 8.2 & \nd  8.3 &  \nd 7.5 & \nd  7.6\\
[0.8pt] 

\multirow{1}{*}{\makecell[l]{\textbf{\methodname{}} \textbf{(Ours)}}}
& ATE [cm]$\downarrow$ &  \nd 6.1 & \win{8.9} &  \nd 7.4 & \win{8.0} & \win{7.4} & \win{7.3} & \win{7.5} \\
\bottomrule
\end{tabularx}
\caption{\textbf{Trajectory evaluation on ScanNet~\cite{Dai2017ScanNet}.}  We achieve state-of-the-art results on trajectory estimation. We refer to qualitative results on this dataset (Fig.~\ref{fig:scannet_qual_169_207}) to highlight the deblurring capabilities of \methodname{}.
 }
\label{tab:render_scannet}
\end{table*}
\begin{table*}
\centering
\scriptsize
\setlength{\tabcolsep}{15.1pt}
\renewcommand{\arraystretch}{1.1}
\begin{tabularx}{\linewidth}{llrrrrrr}
\toprule
Method & Method  & \texttt{f1/desk} & \texttt{f2/xyz} & \texttt{f3/off}  & \texttt{f1/desk2} & \texttt{f1/room} & \textbf{Avg.}\\
\midrule

\multirow{1}{*}{\makecell[l]{MonoGS~\cite{matsuki2024gaussian}}} 
& ATE [cm]$\downarrow$ & \rd  3.8 & \rd  5.2 & \rd  2.9 & \rd  75.7 & \rd  76.6 & \rd 32.84\\
[0.8pt] 

\multirow{1}{*}{\makecell[l]{Splat-SLAM~\cite{sandstrom2024splat}}} 
& ATE [cm]$\downarrow$ & \win{1.6} & \win{0.2} & \win{1.4} & \nd  2.8 & \win{4.2}  & \nd  2.04\\
[0.8pt] 

\multirow{1}{*}{\makecell[l]{\textbf{\methodname{}} \textbf{(Ours)}}}
& ATE [cm]$\downarrow$ & \win{1.6} & \win{0.2} & \nd 1.5 & \win{2.5} & \win{4.2} & \win{2.00} \\
\bottomrule
\end{tabularx}
\vspace{-3pt}
\caption{\textbf{Trajectory evaluation on TUM-RGBD~\cite{sturm2012benchmark}.} As in Table~\ref{tab:render_scannet}, we evaluate the tracking performances of our method. Our method yields similar or better tracking errors than Splat-SLAM and strongly outperforms MonoGS. }
\label{tab:render_tum}
\vspace{-4pt}
\end{table*}
\begin{table}
\vspace{4pt}
\centering
\scriptsize
\setlength{\tabcolsep}{6.0pt}
\renewcommand{\arraystretch}{1.1}
\begin{tabular}{llrrrr}
\toprule
Method & Metric  & \texttt{f1/desk} & \texttt{f2/xyz} & \texttt{f3/off} & \textbf{Avg.}\\
\midrule

\multirow{1}{*}{\makecell[l]{$I^2$-SLAM~\cite{bae20242}}} 
& ATE [cm]$\downarrow$ & \fs 1.6 & \nd 0.3 & \nd 1.9 & \nd 1.3 \\
[0.8pt] 

\multirow{1}{*}{\makecell[l]{\textbf{\methodname{}}}}
& ATE [cm]$\downarrow$ & \fs 1.6 & \fs 0.2 & \fs 1.5 & \fs 1.1 \\
\bottomrule
\end{tabular}
\caption{\textbf{Tracking performances on TUM-RGBD~\cite{sturm2012benchmark},} compared to the only related method that solves a similar problem. Our method achieves more accurate trajectory estimates (other properties cannot be evaluated since their code is not publicly released). }
\label{tab:render_ate_tum}
\end{table}

\begin{table}[t]
    \vspace{-6pt}
    \centering
    \scriptsize
    \setlength{\tabcolsep}{5.1pt}
    \begin{tabular}{llrrr}
    \toprule
    Method & Metric & \texttt{sofa 1} & \texttt{sfm\_bench} & \texttt{plant\_scene\_1} \\
    \midrule
    \multirow{7}{*}{\makecell[l]{\textbf{SplatSLAM}\\ {\cite{sandstrom2024splat}}}} 
    & PSNR$\uparrow$ & \nd 20.50 & \nd 15.44 & \win 28.42\\
    & SSIM$\uparrow$ & \nd 0.66 & \nd 0.46 & \win 0.87\\
    & LPIPS$\downarrow$ & \nd 0.43 \nd & \nd 0.56 & \nd 0.18\\
    & sPSNR$\uparrow$ & \nd 14.09 & \nd 12.07 & \nd 21.74\\
    & sSSIM$\uparrow$ & \nd 0.53 & \nd 0.39 & \nd 0.77\\
    & sLPIPS$\downarrow$ & \nd 0.61 & \nd 0.65 & \nd 0.28\\
    [0.8pt] \hdashline \noalign{\vskip 1pt}
    \multirow{7}{*}{\makecell[l]{\textbf{\methodname{}}\\ \textbf{(Ours)}}} 
    & PSNR$\uparrow$ & \win 26.60 & \win 19.96 & \nd 28.38\\
    & SSIM$\uparrow$ & \win 0.89 & \win 0.71 & \win 0.87\\
    & LPIPS$\downarrow$ & \win 0.18 & \win 0.35 & \win 0.14\\
    & sPSNR$\uparrow$ & \win 15.00 & \win 13.72 & \win 23.00\\
    & sSSIM$\uparrow$ & \win 0.62 & \win 0.55 & \win 0.81\\
    & sLPIPS$\downarrow$ & \win 0.56 & \win 0.51 & \win 0.22\\
    \bottomrule
    \end{tabular}
    \caption{\textbf{Rendering evaluation on \texttt{ETH3D}~\cite{Schops_2019_CVPR}} with synthetically blurred frames. By averaging frames to introduce blur we can evaluate method performance against GT. Our method outperforms Splat-SLAM on both sharp (denoted with "s" prefix) and averaged rendering metrics.}
    \label{tab:eval_ethed_avg}
    \vspace{-8pt}
\end{table}

\boldparagraph{Datasets.}
To evaluate our method, we generate a synthetically blurred Replica dataset made from the scenes used in~\cite{matsuki2024gaussian}. By linearly interpolating the existing trajectory we added 8 sub-poses in between the existing ones. 
We rendered the new frames using Habitat-Sim~\cite{szot2021habitat} and then averaged 36 frames to create a blurry dataset for evaluation.
We also test on real-world data using the TUM-RGBD~\cite{sturm2012benchmark}, ScanNet datasets~\cite{Dai2017ScanNet} and ETH3D~\cite{Schops_2019_CVPR} datasets.

\boldparagraph{Metrics.}
For evaluation on Replica~\cite{straub2019replica}, we report PSNR, SSIM~\cite{wang2004image} and LPIPS~\cite{zhang2018unreasonable} on the rendered keyframe images against the original sharp images (denoted by "s" in front as in Table~\ref{tab:render_replica}).
Thus, we also evaluate the deblurring capabilities.
For evaluation on ScanNet and TUM-RGBD, we evaluate the Absolute Trajectory Error (ATE) \cite{sturm2012benchmark} between the virtual camera poses (as interpolated between two control points) and the ground truth poses used to render the dataset in Replica. For real-world datasets, we evaluate the mid-pose of the two control points against the ground truth.

\boldparagraph{Implementation details.}
We use 9 virtual cameras for the synthetic Replica dataset (as we use the original one and the 8 generated ones) to maintain a good correspondence with the original data while taking advantage of the fact that 9 is a divisor of the 36 averaged frames. 
We use 5 virtual cameras for the real-world data (but compare only to a single ground truth pose as sub-frame ground truth is unavailable).
We keep the same frame-to-frame parameters as in \cite{sandstrom2024splat}. 
Experiments are conducted on an AMD EPYC 7742 processor and NVIDIA GPU GeForce RTX 3090 with 24 GB memory.

\boldparagraph{Results on synthetic data.}
We tested our method on the proposed synthetic Replica dataset with sub-frame ground truth. 
Qualitative results are in Figs.~\ref{fig:office2_qual} and~\ref{fig:office3_qual}. 
These results show that our method successfully models the physical blur formation and renders sharp virtual images. 
Our method outperforms all other methods on all metrics (Table \ref{tab:render_replica}).

\boldparagraph{Results on real data.}
In real-world datasets, the absence of proper ground truth prevents testing the sharp frames directly. 
A recent method solving a similar problem has been introduced in $I^2$-SLAM~\cite{bae20242}, but sharp frames are manually selected, and metrics are evaluated solely on these frames, potentially introducing bias. 
To date, there is no available information about the specific frames used, nor has the code been shared, making comparisons challenging. 
Only trajectory evaluation on TUM-RGBD could be compared, shown in Table~\ref{tab:render_ate_tum}.
Our method outperforms $I^2$-SLAM in this setting.
Additionally, we present evaluation on ScanNet and TUM-RGBD datasets in Tables~\ref{tab:render_scannet} and~\ref{tab:render_tum}, evaluating the trajectory error against other baselines. Our method outperforms all other methods on trajectory error (ATE). 

\boldparagraph{Test on Blurry Real-World Dataset.}
To further assess our model, we conducted experiments on the TUM-RGBD dataset~\cite{sturm2012benchmark} and ETH3D~\cite{Schops_2019_CVPR}, where we increased the motion blur by averaging 2-5 frames. We matched the number of virtual cameras to the number of averaged frames, allowing us to pair each virtual image with a high-quality ground-truth frame. This setup enables us to showcase the effectiveness of our method in a different scenario.
We managed to surpass all other metrics, demonstrating the overall effectiveness of our approach. We achieve better outcomes, especially for sharp virtual images, and observe a significant improvement in the ATE. The results can be seen in Table \ref{tab:eval_ethed_avg}. More experiments and qualitative results can be found in the supplementary material.

\boldparagraph{Runtime.} 
We tested the runtime of our method on ScanNet scene \texttt{0106\_00}, where it is 0.36 fps. 
Consequently, in terms of virtual sub-frame images, the runtime is 1.82 fps. 
This evaluation includes only the online process and not the final offline refinement at the end. 

\boldparagraph{Limitations.}
Our method does not model and recover motion blur from moving objects. This requires further decomposition and is an interesting future work.
In practice, not all video frames are blurry, and the high computational overhead for deblurring could be reduced by dynamically adapting the generated number $M$ of virtual views.
%
\section{Conclusions}

We presented \methodname{}, a robust RGB SLAM pipeline to recover sharp reconstructions from blurry inputs. 
Our pipeline takes advantage of sub-frame trajectory modeling, online loop closure, and global BA to achieve dense and accurate trajectories. 
Moreover, we model the physical image formation process of motion-blurred images and ensure sharp, sub-frame, and precise mapping thanks to the monocular depth estimator and online deformation of Gaussians. 
\methodname{} applies these techniques in the context of scene deblurring for the first time.
Experiments show that our method outperforms existing methods.

{
    \small
    \bibliographystyle{ieeenat_fullname}
    \bibliography{main}
}
\maketitlesupplementary

\begin{figure*}
\centering
 \includegraphics[width=\linewidth]{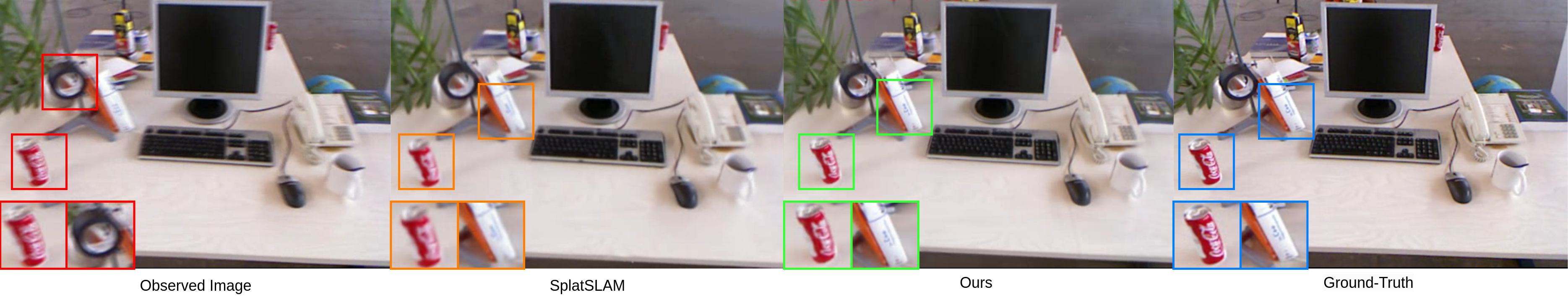}\\[-0.3em]
\caption{\textbf{Qualitative results on averaged real-world TUM-RGBD data (\texttt{f2/xyz})~\cite{sturm2012benchmark}}, in which we averaged 5 frames. Given the blurry input frames, we can track the trajectory with sub-frame precision and estimate sharp maps by directly modeling the camera motion blur.} 
\label{fig:tum_avg}
\end{figure*}

\section{Method Details}

\boldparagraph{Hyperparameters.}
We use 150 iterations for mapping and 200 iterations for the frame-to-model tracking. 
We set $\lambda = 0.8$, $\lambda_{SSIM} = 0.2$, $\lambda_{reg} = 10$.
The learning rate for the rotational relative pose is set to $3e-3$ and $1e-3$. We keep the early stopping in the frame-to-model tracking when the gradient goes under $1e-4$, as introduced in \cite{matsuki2024gaussian} for faster tracking. For the other hyperparameters please refer to \cite{matsuki2024gaussian,zhang2024glorie,sandstrom2024splat}

\boldparagraph{Keyframes Management.}
We use the keyframe window as in MonoGS~\cite{matsuki2024gaussian}. 
The keyframe selection is based on the covisibility of the Gaussians. The covisibility between two keyframes $i$, $j$, is defined using the Intersection over Union (IoU) and Overlap Coefficient (OC):
\begin{equation}
    \text{IoU}_{\text{cov}}(i, j) = \frac{|\mathcal{G}_v^i \cap \mathcal{G}_v^j|}{|\mathcal{G}_v^i \cup \mathcal{G}_v^j|}, \label{eq:iou_cov}
\end{equation}
\begin{equation}
    \text{OC}_{\text{cov}}(i, j) = \frac{|\mathcal{G}_v^i \cap \mathcal{G}_v^j|}{\min(|\mathcal{G}_v^i|, |\mathcal{G}_v^j|)}, \label{eq:oc_cov}
\end{equation}
where $\mathcal{G}_v^i$ are the Gaussians visible in keyframe $i$. A Gaussian is defined visible in a rasterization if the ray’s accumulated transmittance $\alpha$ has not yet reached $0.5$. 

A keyframe $i$ is added to the keyframe window KFs if, given the last keyframe $j$, $\text{IoU}_{\text{cov}} (i,j) < k_{\text{fcov}}$ or if the relative translation $t_{ij} > k_{fm} \hat{D}_i$, where $\hat{D}_i$ is the median depth of the mid pose of frame $i$. For Replica we use $k_{\text{fcov}} = 0.95$, $k_{\text{fm}} = 0.04$ and for TUM-RGBD~\cite{sturm2012benchmark} and ScanNet~\cite{Dai2017ScanNet} datasets, $k_{\text{fcov}} = 0.90$, $k_{\text{fm}} = 0.08$. The registered keyframe $j$ in $\text{KFs}$ is removed if $\text{OC}_{\text{cov}}(i, j) < k_{\text{fc}}$, where keyframe $i$ is the latest added keyframe. For all datasets, the cutoff is set to $k_{\text{fc}} = 0.3$. The size of the keyframe window is set to $|\text{KFs}| = 10$ for Replica and $|\text{KFs}| = 8$ for TUM-RGBD and ScanNet.

\boldparagraph{Pruning and Densification.}
The Pruning and Densification are done similarly to MonoGS~\cite{matsuki2024gaussian}. Pruning in the system is guided by visibility criteria. Newly inserted Gaussians from the last three keyframes are discarded if they are not observed by at least three other frames within the keyframe window, but this process only occurs when the keyframe window reaches its capacity. Additionally, for every 150 mapping iterations, a global pruning step removes Gaussians with opacity below 0.7 or those with excessively large 2D projections. Densification follows the method outlined in \cite{kerbl20233d} and is similarly performed every 150 mapping iteration. To avoid forgetting Gaussian, while mapping the current window, we select a few indices of past keyframes and add them to the mapping optimization to preserve rendering quality and geometry on the whole scene.

\boldparagraph{Refinement.}
Following the method outlined in \cite{sandstrom2024splat}, online global BA is conducted every 20 keyframes.
We perform both the GlORIE-SLAM~\cite{zhang2024glorie} trajectory refinement and MonoGS~\cite{matsuki2024gaussian} color refinement at the end of the online optimization. We keep 26000 as the number of refinement optimizations as in MonoGS.

\boldparagraph{Monocular Depth Estimator.}
We used OmniDepth~\cite{eftekhar2021omnidata} as a Monocular Depth Estimator as employed in \cite{sandstrom2024splat} to fill the depth map where the multi-view is invalid. We have tried Depth Anything v2~\cite{yang2024depthv2}, but in our blurry setting, we found OmniDepth to be more robust (Table \ref{tab:mono_depth}).

\begin{table}
\centering
\footnotesize
\setlength{\tabcolsep}{15.5pt}
\begin{tabularx}{\columnwidth}{llr}
\toprule
Method & Metric & \texttt{office0} \\
\midrule
\multirow{3}{*}{\makecell[l]{\textbf{\methodname{}}\\ \textbf{(DepthAnything V2~\cite{yang2024depthv2})}}} 
& sPSNR$\uparrow$ & \nd 30.06 \\
& sSSIM$\uparrow$ & \nd 0.87 \\
& sLPIPS$\downarrow$ & \nd 0.25 \\
[0.8pt] \hdashline \noalign{\vskip 1pt}
\multirow{3}{*}{\makecell[l]{\textbf{\methodname{}}\\ \textbf{(Omnidepth~\cite{eftekhar2021omnidata})}}} 
& sPSNR$\uparrow$ & \win{30.40} \\
& sSSIM$\uparrow$ & \win{0.87}  \\
& sLPIPS$\downarrow$ & \win{0.24} \\
\bottomrule
\end{tabularx}
\caption{\textbf{\textbf{Monocular Depth Estimator evaluation on Replica~\cite{straub2019replica}}}. The results are almost equal, showing that there is not much difference between the two Monocular Depth Estimators. We selected OmniDepth~\cite{eftekhar2021omnidata} being the most consistent, even in blurry frames.}
\label{tab:mono_depth}
\end{table}

\boldparagraph{Details on the Scale and Shift estimation.}
Following \cite{sandstrom2024splat}, the scale $\theta_i$ and shift $\gamma_i$ are initially estimated using least squares fitting using:
\begin{equation}
  \label{eq:scale_shift_eq}
  \left\{\theta_i, \gamma_i\right\}=\underset{\theta, \gamma}{\arg \min } \sum_{(u, v)}\left(\left(\theta\left(1 / D_i^{\text{mono }}\right)+\gamma\right)-d_i^l\right)^2
\end{equation}

\boldparagraph{Disparity Maps Details.}
For a given disparity map $d_i$ (separated into low and high error parts $\{d_i^l , d_i^h\}$) for frame $i$, we denote the corresponding depth $\tilde{D}_i = 1 / d_i$. Similar to \cite{sandstrom2024splat}, pixel correspondences $(u, v)$ and $(\hat{u}, \hat{v})$ between keyframes $i$ and $j$ are established by warping $(u, v)$ into frame $j$ with depth $\tilde{D}_i$ as
\begin{equation}
  p_i=\omega_i \tilde{D}_i(u, v) K^{-1}[u, v, 1]^T, \,\, [\hat{u}, \hat{v}, 1]^T \!\!\propto K \omega_j^{-1}\!\left[p_i, 1\right]^T
\end{equation}
The 3D point corresponding to $(\hat{u}, \hat{v})$ is computed from the depth at $(\hat{u}, \hat{v})$ as
\begin{equation}
  p_j=\omega_j \tilde{D}_j(\hat{u}, \hat{v}) K^{-1}[\hat{u}, \hat{v}, 1]^T 
\end{equation}
If the L2 distance between $p_i$ and $p_j$ is smaller than a threshold, the depth $\tilde{D}_i(u, v)$ is consistent between $i$ and $j$. By looping over all keyframes except $i$, the global two-view consistency $n_i$ can be computed for frame $i$ as
\begin{equation}
  n_i(u, v)=\!\!\!\sum_{\substack{k \in \mathrm{KFs}, k \neq i}} \!\!\! \mathbbm{1}\left(\left\|p_i-p_k\right\|_2<\eta \cdot \operatorname{avg}\big(\tilde{D}_i\big)\right)
\end{equation}
Here, $\mathbbm{1}(.)$ is the indicator function, $\eta \in \mathbb{R}_{\geq 0}$ is a hyperparameter, and $n_i$ is the total two-view consistency for pixel $(u, v)$ in keyframe $i$. 
Moreover, $\tilde{D}_i(u,v)$ is valid if $n_i$ is larger than a threshold.

\section{Additional Experimental Results}

\boldparagraph{Effect on the number of virtual views.}
We analyze the impact of varying the number of virtual cameras in \texttt{office0} to evaluate its influence on the rendering performance. As shown in Table~\ref{tab:virtuals_ablation}, our results reveal that an increasing the number of virtual cameras consistently improves rendering quality. This highlights the significance of optimizing the number of virtual cameras to enhance the overall performance of the system.
\begin{table}
    \centering
    \scriptsize
    \setlength{\tabcolsep}{9.75pt}
    \begin{tabularx}{\columnwidth}{llrrr}
    \toprule
    Method & Metric & $M=3$  & $M=6$ & $M=9$ \\
    \midrule
    \multirow{3}{*}{\makecell[l]{\textbf{\methodname{}}\\ \textbf{(Ours)}}} 
    & sPSNR$\uparrow$ & \rd 28.90 & \nd 30.11 & \win{30.40} \\
    & sSSIM$\uparrow$ & \rd 0.85 & \nd 0.87 & \win{0.87}  \\
    & sLPIPS$\downarrow$ & \rd 0.29 & \nd 0.25 & \win{0.24} \\
    \bottomrule
    \end{tabularx}
    \caption{\textbf{Ablation on the number of virtual cameras on \texttt{office0}~\cite{straub2019replica}}. We observe that the rendering metrics increase with the number of virtual cameras used. This indicates that more virtual cams make the blur modeling more accurate and lead to sharp renderings.}
    \label{tab:virtuals_ablation}
\end{table}

\boldparagraph{Ablation on features.}
In our study, we conducted a comprehensive ablation analysis on key components of our model to demonstrate their individual contributions to overall performance. The results, detailed in Table~\ref{tab:ablation}, clearly indicate that our full method, incorporating physical blur modeling, frame-to-frame tracking, Gaussian deformations, and final refinement, surpasses all variants lacking one or more of these elements in rendering quality. This emphasizes the integral role each component plays in enhancing our method's rendering capabilities, validating our approach of integrating these techniques for optimal results. 
\begin{table}
    \centering
    \scriptsize
    \setlength{\tabcolsep}{1.3pt}
    \begin{tabularx}{\columnwidth}{lcccclr}
    \toprule
    Method & \makecell{Physical\\Blur Model} & \makecell{Frame-to-\\Frame} & \makecell{Gaussian\\ Deformation} & \makecell{Refine-\\ment} & Metric & \texttt{office0} \\
    \midrule
    \multirow{3}{*}{\makecell[l]{\textbf{MonoGS}\\ \cite{matsuki2024gaussian}}} 
    & & & & &sPSNR$\uparrow$ & 25.30 \\
    & \xmark & \xmark & \xmark & \cmark &sSSIM$\uparrow$ & 0.82 \\
    & & & & &sLPIPS$\downarrow$ & 0.43 \\
    [0.8pt] \hdashline \noalign{\vskip 1pt}
    \multirow{3}{*}{\makecell[l]{\textbf{MonoGS}\\ + Deblurring}} 
    & & & & & sPSNR$\uparrow$ & \rd 25.91 \\
    & \cmark & \xmark & \xmark & \xmark &sSSIM$\uparrow$ & \rd 0.83 \\
    & & & & &sLPIPS$\downarrow$ & \rd 0.40 \\
    [0.8pt] \hdashline \noalign{\vskip 1pt}
    \multirow{3}{*}{\makecell[l]{\textbf{\methodname{}}\\ No Deformation \\ No Refinement}} 
    & & & & &sPSNR$\uparrow$ & \nd 28.23 \\
    & \cmark & \cmark & \xmark & \xmark & sSSIM$\uparrow$ & \nd 0.84  \\
    & & & & & sLPIPS$\downarrow$ & \nd 0.33 \\
    [0.8pt] \hdashline \noalign{\vskip 1pt}
    \multirow{3}{*}{\makecell[l]{\textbf{\methodname{}}\\ \textbf{(Ours)}}} 
    & & & & &sPSNR$\uparrow$ & \win{30.40} \\
    & \cmark & \cmark & \cmark & \cmark & sSSIM$\uparrow$ & \win{0.87}  \\
    & & & & &sLPIPS$\downarrow$ & \win{0.24} \\
    \bottomrule
    \end{tabularx}
    \caption{\textbf{Ablation on the different features of our method on \texttt{office0}~\cite{straub2019replica}}. This table showcases the individual contributions of various features in our method, demonstrating how each component enhances overall performance and justifying our design choice.}
    \label{tab:ablation}
\end{table}

\boldparagraph{Test on Blurry Real-World Dataset.}
We averaged 5 frames for the TUM-RGBD dataset and 2 frames for the ETH3D dataset.
On Table \ref{tab:eval_tum_avg}, we show that \methodname{} can outperform almost all the metrics. More specifically, the better sPSNR, sSSIM, sLPIPS denote that our methods successfully learn a sharper map.
An example result in comparison with SplatSlam~\cite{sandstrom2024splat} is depicted in Figure~\ref{fig:tum_avg}.
\begin{table}
    \centering
    \scriptsize
    \setlength{\tabcolsep}{10.3pt}
    \begin{tabularx}{\columnwidth}{llrrr}
    \toprule
    Method & Metric & Avg 3fr & Avg 5fr & Avg 7fr\\
    \midrule
    \multirow{7}{*}{\makecell[l]{\textbf{SplatSLAM}\\ \cite{sandstrom2024splat}}} 
    & PSNR$\uparrow$ & \fs 27.64 &  \fs 27.32 & \fs 27.11\\
    & SSIM$\uparrow$ & \nd 0.88 & \nd 0.87 & \fs 0.88\\
    & LPIPS$\downarrow$ & \nd 0.14 & \nd 0.17 & \fs 0.17\\
    & sPSNR$\uparrow$ & \nd 20.44 & \nd 17.61 & \nd 20.18\\
    & sSSIM$\uparrow$ & \nd  \nd 0.68 & \nd 0.59 & \nd 0.66\\
    & sLPIPS$\downarrow$ & \nd 0.22 & \nd 0.30 & \nd 0.30\\
    & ATE$\downarrow$ & \nd 21.8 & \nd 21.8 & \nd 21.8\\
    [0.8pt] \hdashline \noalign{\vskip 1pt}
    \multirow{7}{*}{\makecell[l]{\textbf{\methodname{}}\\ \textbf{(Ours)}}} 
    & PSNR$\uparrow$ & \nd 27.26 & \nd 27.00 & \nd 25.90\\
    & SSIM$\uparrow$ & \fs 0.90 & \fs 0.89 & \nd 0.87\\
    & LPIPS$\downarrow$ & \fs 0.11 & \fs 0.14 & \fs 0.17\\
    & sPSNR$\uparrow$ & \fs 20.70 & \fs 20.70 & \fs 20.37\\
    & sSSIM$\uparrow$ & \fs 0.70 & \fs 0.70 & \fs 0.68\\
    & sLPIPS$\downarrow$ & \fs 0.18 & \fs 0.21 & \fs 0.26\\
    & ATE$\downarrow$ & \fs 0.2 & \fs 0.2 & \fs 0.2\\
    \bottomrule
    \end{tabularx}
    \caption{\textbf{Rendering evaluation on custom \texttt{f2/xyz} \cite{sturm2012benchmark}} with averaged frames. By averaging some frames to introduce blur, we show that our method outperforms others in rendering metrics for sharp images (denoted with the "s" prefix) as well as better ATE. \methodname{} achieves comparable results for the averaged renderings metrics.}
    \label{tab:eval_tum_avg}
\end{table}
\begin{table}[t]
\centering
\scriptsize
\setlength{\tabcolsep}{17.9pt}
\begin{tabularx}{\linewidth}{llr}
\toprule
Method & Metric & \texttt{106\_00} \\
\midrule
\multirow{2}{*}{\makecell[l]{\textbf{\methodname{} Light}}}
& FPS [frame/s]$\uparrow$ & \win 0.65 \\
& PSNR$\uparrow$ & \nd 22.90\\
[0.8pt] \hdashline \noalign{\vskip 1pt}
\multirow{2}{*}{\makecell[l]{\textbf{\methodname{}}}}
& FPS [frame/s]$\uparrow$ & \nd 0.36 \\
& PSNR$\uparrow$ & \win 23.06 \\
\bottomrule
\end{tabularx}
\vspace{-0.2em}
\caption{\textbf{Speed performances on ScanNet~\cite{Dai2017ScanNet},} compared to a lighter version of our model with 120 iterations for mapping and 20 for frame-to-model tracking.
}
\label{tab:speed_ablation}
\vspace{-8pt}
\end{table}

\end{document}